%% file: main.tex
\documentclass[conference]{IEEEtran}

\usepackage{amsthm}
\usepackage{cite}
\usepackage{amsmath}
\usepackage{amssymb}
\usepackage{graphicx}
\usepackage{booktabs}
\usepackage{subfigure}
\usepackage{textcomp}
\newtheorem{theorem}{Theorem}
\newtheorem{lemma}[theorem]{Lemma}
\usepackage{algorithm}
\usepackage{algorithmicx}
\usepackage{balance}
\usepackage{algpseudocode}
\usepackage[misc,geometry]{ifsym}

\newcommand{\ie}[0]{\textit{i.e.}}

\begin{document}
%
\title{Multi-view Graph Embedding with Hub Detection for Brain Network Analysis}

\author{\IEEEauthorblockN{Guixiang Ma$^{\ast}$, Chun-Ta Lu$^{\ast}$, Lifang He$^{\ddag}$,  Philip S. Yu$^{\P,\ast}$, Ann B. Ragin$^{\dag}$}
\IEEEauthorblockA{$^{\ast}$University of Illinois at Chicago, Chicago, IL, USA\\
   $^{\P}$Shanghai Institute for Advanced Communication and Data Science, Fudan University, Shanghai, China\\
   $^{\ddag}$Department of Healthcare Policy and Research, Cornell University\\
   $^{\dag}$Northwestern University, Chicago, IL, USA\\ \{gma4, clu29, psyu\}@uic.edu, lifanghescut@gmail.com, ann-ragin@northwestern.edu}}

\maketitle

\begin{abstract}
Multi-view graph embedding has become a widely studied problem in the area of graph learning. Most of the existing works on multi-view graph embedding aim to find a shared common node embedding across all the views of the graph by combining the different views in a specific way. Hub detection, as another essential topic in graph mining has also drawn extensive attentions in recent years, especially in the context of brain network analysis. Both the graph embedding and hub detection relate to the node clustering structure of graphs. The multi-view graph embedding usually implies the node clustering structure of the graph based on the multiple views, while the hubs are the boundary-spanning nodes across different node clusters in the graph and thus may potentially influence the clustering structure of the graph. However, none of the existing works in multi-view graph embedding considered the hubs when learning the multi-view embeddings. In this paper, we propose to incorporate the hub detection task into the multi-view graph embedding framework so that the two tasks could benefit each other. Specifically, we propose an auto-weighted framework of Multi-view Graph Embedding with Hub Detection (MVGE-HD) for brain network analysis. The MVGE-HD framework learns a unified graph embedding across all the views while reducing the potential influence of the hubs on blurring the boundaries between node clusters in the graph, thus leading to a clear and discriminative node clustering structure for the graph. We apply MVGE-HD on two real multi-view brain network datasets (\ie, HIV and Bipolar). The experimental results demonstrate the superior performance of the proposed framework in brain network analysis for clinical investigation and application. 
\end{abstract}


%
\IEEEpeerreviewmaketitle

\input{intro_01}

\input{problem_02}

\input{method_03}

\input{optimize_04}
\input{experiment_05}
\input{relatedwork_06}
\input{conclusion_07}
 



\bibliographystyle{IEEEtran}

\end{document}

%% file: intro_01.tex
\section{Introduction} \label{sec_intro}
Recent years have witnessed an explosion of data in the form of graph representations. These data comes with a set of nodes and links between the nodes, for example the social networks with nodes representing users and links representing relationships among the users, and the brain networks with brain regions as nodes and the correlations among different regions as links. With the advanced capabilities for data acquisition, the links can usually be constructed from multiple sources or views of the data, which is usually called multi-view graph data. For instance, brain networks can be derived from fMRI (functional magnetic resonance imaging) and DTI (diffusion tensor imaging), which are two major neuroimaging techniques for brain data acquisition in neuroscience research and clinical applications. The fMRI brain networks reflect the correlations of different brain regions in functional activity, while the DTI networks encode the information of structural connections (\textit{i.e.} white matter fiber paths) between different brain regions. Thus these two kinds of networks can serve as two views of the connectivity for brain network data \cite{ma2017multi}.

Multi-view graph embedding, as a hot topic in  multi-view graph learning, has drawn extensive attentions in the past decade. Most of the existing works in multi-view graph embedding aim to combine the information from all the views and obtain a lower dimensional but better feature representation of the nodes for the spectral clustering problem. For example, in \cite{kumar2011co}, a co-regularized multi-view spectral clustering method is proposed to find a consistent clustering across the multiple views. In \cite{papalexakis2013more}, two solutions based on minimum description length and tensor decomposition principles are proposed for graph clustering across multiple views. A multi-modal spectral clustering algorithm is presented in \cite{cai2011heterogeneous} to learn a commonly shared graph Laplacian matrix by unifying different views. In \cite{huang2012affinity}, an affinity aggregation spectral clustering algorithm is proposed, which seeks for an optimal combination of affinity matrices for the spectral clustering across multiple views. In \cite{li2015large}, a large-scale multi-view spectral clustering approach is proposed using local manifold fusion to integrate heterogeneous features of graphs.

Although these works introduced above can be used to obtain the graph embeddings from multiple views, none of them has considered the hubs when learning the multi-view graph embedding, making them less capable for the scenarios where hubs are also important for the clustering of nodes in graphs. The ``hubs" refer to the bridging nodes that connect to different groups of nodes in a graph. For example, in a brain network, the hubs help bridge different groups of brain regions\cite{van2013network}, while in a social network, the hubs are known as ``structural hole spanners"\cite{he2016joint}, which refer to the users bridging different communities. The hubs in both of the two scenarios can potentially influence the node clustering structure of the network, as they are the boundary-spanning nodes across different clusters and their neighbors usually spread out in different clusters. Therefore, hubs should be taken into account in the multi-view graph embedding learning process for achieving a clear and discriminative node clustering structure for the graph. Specifically, in neuroscience studies, the hubs of brain networks have been proven to be more biologically costly due to higher blood flow or connection distances, thus they tend to be more vulnerable to brain injuries \cite{crossley2014hubs}. As a result, the hubs will differ in the brain networks of normal people and those of the subjects with neurological disorder, which means the corresponding brain network embeddings of normal people and disordered subjects also tend to be different. Therefore, if we could consider the hubs when learning multi-view graph embeddings of brain networks, the resulted embeddings will be useful for distinguishing brain disordered subjects from normal controls.

In this paper, we focus on jointly learning the multi-view graph embeddings and hubs for brain network analysis. There are three main challenges that must be addressed for this problem:
\begin{itemize}
\item As the task of multi-view graph embedding and the task of multi-view hub detection are naturally twisted, how to provide a joint learning framework such that both tasks can be solved at the same time and help improve the overall performance. 

\item It is often assumed that each individual view captures the partial information but they all admit the same underlying structure of the data. How to leverage the multi-view graph data for obtaining a good unified graph embedding across all the views?
 
\item How to decide the importance of each view of the data when combining them for the multi-view learning task?
\end{itemize}
  
To address the above challenges, we propose an auto-weighted multi-view graph embedding with hub detection (MVGE-HD) framework. Our contributions can be summarized as follows:
\begin{itemize}
\item To the best of our knowledge, this is the first work to solve the problem of multi-view graph embedding with hub detection.

\item The proposed MVGE-HD framework can jointly learn the multi-view graph embeddings and identify the hubs, instead of separating them into different steps. By considering the hubs, the obtained embeddings will reflect a clearer node clustering structure of the graph, which can better facilitate the further analysis of the graph. 

\item Our framework can automatically tune the importance of each view for the multi-view graph embedding with hub detection, avoiding the problem that might be caused otherwise by different parameter settings and thus having good generalization ability.  

\item We apply the MVGE-HD framework on two real brain network datasets (HIV and Bipolar) to investigate the multi-view brain region clustering structure and the hubs in brain networks for neurological disorder analysis, as a topic discussed for the first time in the literature of neuroscience study as well. The experimental results show the effectiveness of MVGE-HD for multi-view brain network analysis.

\end{itemize}

The rest of this paper is organized as follows. The problem formulation and some preliminary knowledge are given in the next section. Then we present the details of the proposed MVGE-HD framework in Section \ref{sec_method} and \ref{sec_opt}. The experimental results and analysis are shown in Section \ref{sec:exp}. Related works are discussed in Section \ref{sec:related} and the conclusions in Section \ref{sec_conclusion}.

%% file: problem_02.tex
\section{Preliminaries} \label{sec_problem}
In this section, we introduce some notations and terminologies that we will use in this paper. Then we establish some definitions and formulate the problem formally.
\begin{figure}[t]
\centering
    \begin{minipage}[l]{0.75\linewidth}
      \centering
      \includegraphics[width=\linewidth]{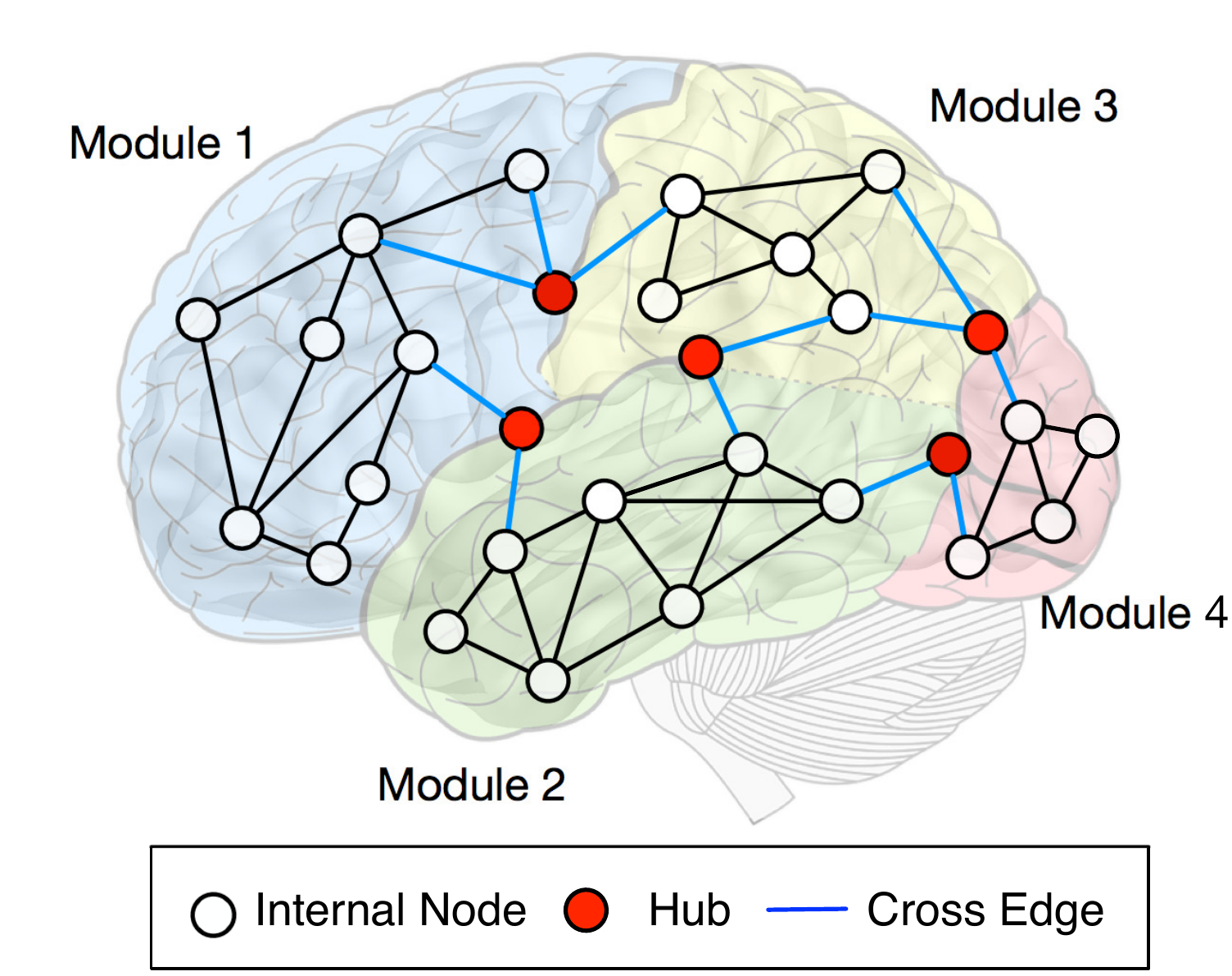}
    \end{minipage}
  \caption{An brain network example with four modules and five hubs }\label{fig:brain_net}
\end{figure}

\textbf{Notations.} Vectors are denoted by boldface lowercase letters, and matrices are denoted by boldface capital letters. An element of a vector $\mathbf{x}$ is denoted by $x_i$, and an element of a matrix $\mathbf{X}$ is denoted by $x_{ij}$. For a matrix $\mathbf{X}  \in \mathbb{R}^{n \times m}$, its $i$-th row and $j$-th column are denoted by $\mathbf{x}^{i}$ and $\mathbf{x}_{j}$, respectively. The Frobenius norm of $\mathbf{X}$ is defined as $\left \| \mathbf{X} \right \|_F = \sqrt{\sum_{i=1}^{n} \|\mathbf{x}^{i} \|_2^2}$, and the $\ell_{2,1}$ norm of $\mathbf{M}$ is defined as $\left \| \mathbf{M} \right \|_{2,1} = \sum_{i=1}^{n} \left \| \mathbf{m}^{i}  \right\|_2$. For any vector $\mathbf{x} \in \mathbb{R}^n$, $Diag(\mathbf{x}) \in \mathbb{R}^{n \times n}$ is the diagonal matrix whose diagonal elements are $x_i$. $\mathbf{I}_n$ denotes an identity matrix with size $n$. We denote an undirected graph with $m$ views as $G=(V, {E_{(1)},E_{(2)}, \cdots, E_{(m)}})$, where $V$ is the set of nodes and $E_{(i)} \subset V \times V $ is the set of edges from view $i$ of $G$. We denote the affinity matrices of the multi-view graph $G$ by $A =\{\mathbf{A}_{(1)},\mathbf{A}_{(2)},\cdots ,\mathbf{A}_{(m)}\}$, where $\mathbf{A}_{(i)} \in \mathbb{R}^{n \times n}$ is the weighted affinity matrix in view $i$, and its entry denotes the pairwise affinity between nodes of $G$ in view $i$.

We assume $\mathbf{F}\in\mathbf{R}^{n\times k}$ is an embedding of $G$, and then the $i$-th row vector of $\mathbf{F}$ (\ie, $\mathbf{f}^i$) represents the embedding of node $i$. We call $k$ the dimension of the embedding $\mathbf{F}$. If we run k-means algorithm on the set of row vectors of $\mathbf{F}$ and set the number of clusters as $k$, we will get a clustering assignment of the $n$ nodes into $k$ clusters. Thus an embedding of a graph usually implies its node clustering structure. We assume the $k$ clusters are represented by $C = \{C_1,\cdots, C_k\}$, with $V = C_1 \cup \cdots \cup C_k$ and $C_i \cap C_j = \varnothing$ for every pair $i, j$ with $i \neq j$. Based on these assumptions, we give the following definitions. 

\noindent \textbf{Definition 1.} \textbf{(Internal Node)}
For any node $v_i\in C_x$, if all the nodes that $v_i$ have connections with belong to the same cluster $C_x$, node $v_i$ is called an internal node.  \\
\noindent \textbf{Definition 2.} \textbf{(Hub)} For any node $v_i\in C_x$, if there exists some neighboring node $v_j \in C_y(x \neq y)$, node $v_i$ is called a hub.  \\
\noindent \textbf{Definition 3.} \textbf{(Cross Edge)} For any edge $e_{ij} = (v_i, v_j) \in E$, if $v_i \in C_x$ and $v_j \in C_y(x \neq y)$, edge $e_{ij}$ is called a cross edge. 

Fig.~\ref{fig:brain_net} shows an example of a brain network with four modules and five hubs. Note that in brain networks, the clusters of brain regions are often called ``modules". In some works of brain hub analysis, the hubs shown in Fig.~\ref{fig:brain_net} are called ``connector hubs" while another kind of hubs are called ``provincial hubs"~\cite{van2013network}, which refer to the internal node with high centrality within a module. In this paper, the hubs we considered are the ``connector hubs" stated in those works.

%% file: method_03.tex
\section{Methodology} \label{sec_method}
In this section, we first present the proposed approach for multi-view graph embedding with hub detection. Then we derive the auto-weighted framework for the proposed approach.

\subsection{Multi-view Graph Embedding with Hub Detection}\label{sec_method_A}
Graph embedding, as an important tool in topological graph theory, has been widely studied for graph data analysis \cite{belkin2001laplacian,fu2012graph,yan2007graph}. In the literature of graph embedding, hubs are seldom considered along with the embedding learning. However, in many graph learning scenarios, hubs play an important role for node clustering or graph embedding analysis. As shown in Figure 1, the hubs are those boundary-spanning nodes across different clusters in the graph, and their neighbors naturally occur in different clusters, and thus the hubs may blur the boundary between clusters. If we want to obtain a graph embedding that can encode a clear node clustering structure, it is crucial to enable the graph embedding approach to have the discriminative ability for such boundary-spanning nodes, \ie, the hubs, and thus encoding only characterizing internal nodes in the graph. To solve this problem, the $\ell_{2,1}$-norm penalty is introduced to the context of node clustering and has been proven to be an effective strategy for dealing with the boundary-spanning nodes and improving the node clustering~\cite{he2016joint, ma2016multi}. In this paper, we employ the similar strategy and incorporate it into our multi-view graph embedding and hub detection framework.

To derive our multi-view framework, we first formulate the problem of single-view graph embedding with hub detection. Given the affinity matrix $\mathbf{A}_{(v)}$ and the diagonal matrix $\mathbf{D}_{(v)}$ with ${d_{(v)}}_{ii} = \sum_{j=1}^{n} a_{ij}$ for view $v$ of the graph $G$, we intend to obtain a graph embedding $\mathbf{F}_{(v)} \in \mathbf{R}^{n \times k}$. Based on the analysis in \cite{ma2016multi}, the value of $\mathbf{f}_{(v)}^i$ at node $v_i$ can be formulated as the weighted average of $\mathbf{f}_{(v)}^i$ at neighbors of $v_i$, where the weights are proportional to the edge weights in adjacency matrix $\mathbf{A}_{(v)}$, thus we can have the following objective function
\begin{align}
\min_{\mathbf{F}_{(v)}} ~& \left \| \mathbf{F}_{(v)} - \mathbf{D}_{(v)}^{-1} \mathbf{A}_{(v)}\mathbf{F}_{(v)} \right \|_{F}^2
\end{align}

As discussed above, we need to make the embedding matrix $\mathbf{F}_{(v)}$ have discriminative ability for the hubs for inducing a clearer node clustering structure of $G$. Based on \cite{he2016joint} and \cite{ma2016multi}, we apply the $\ell_{2,1}$-norm penalty and orthogonality constraints to promote the row-wise sparsity, so as to discriminate the hubs and encode only characterizing internal nodes. Then the problem of graph embedding with hub detection on single view becomes
\begin{align}
\min_{\mathbf{F}_{(v)}} ~& \left \| \mathbf{F}_{(v)} - \mathbf{D}_{(v)}^{-1} \mathbf{A}_{(v)}\mathbf{F}_{(v)} \right \|_{2,1}\nonumber \\
\text{s.t.} & ~~ \mathbf{F}_{(v)}^{\mathrm{T}}\mathbf{F}_{(v)} = \mathbf{I}_{k}
\label{eq: single_mvge}
\end{align}

For the multi-view graph learning task, we consider combining the information from the multiple views of graph $G$ and obtaining a unified graph embedding across all the views, which can better encode the embedding structure while considering the multi-view hubs as well. To achieve this goal, we propose to use the weighted combination of the graph embedding from each view, and we formulate it as follows.

We assume the unified embedding matrix across all the views of graph $G$ is represented by $\mathbf{F}\in\mathbf{R}^{n\times k}$, where $k$ is the dimension of the row vectors. Then the multi-view graph embedding with hub detection can be formulated as the following problem
\begin{align}
\min_{\mathbf{F}} ~& \sum_{v=1}^{m} \alpha_{(v)}\left \| \mathbf{F} - \mathbf{D}_{(v)}^{-1} \mathbf{A}_{(v)}\mathbf{F} \right \|_{2,1} \nonumber \\
\text{s.t.} & ~~ \mathbf{F}^{\mathrm{T}}\mathbf{F} = \mathbf{I}_{k}
\label{eq: mvge}
\end{align}
where $\alpha_{(v)}$ is the weight parameter for view $v$. Note that here the value of the weight parameter $\alpha_{(v)}$ is decided by an auto-tuning procedure, which will be introduced later in Section \ref{sec_framework}.   

As the above minimization problem involving $\ell_{2,1}$ norm is nontrivial to solve directly, we further derive Eq.~(\ref{eq: mvge}) based on the following lemma \cite{he20122}. 

\begin{lemma} \label{lemma_L21norm}
Let $\phi(.)$ be a function satisfying the conditions: $x \rightarrow \phi(x)$ is convex on $R$; $x \rightarrow \phi(\sqrt{x})$ is convex on $R_{+}$; $\phi(x) = \phi(-x), \forall x \in R$; $\phi(x)$ is $C^{1}$ on $R$; $\phi^{\prime\prime}(0^{+}) \geq 0$, $\underset{x \to \infty}{\lim} \phi(x)/x^2 = 0$. Then for a fixed $\| \mathbf{u}^i \|_2$, there exists a dual potential function $\varphi(.)$, such that
\begin{equation}
    \phi(\| \mathbf{u}^i \|_2) = \underset{p\in R}{\inf} \{ p \| \mathbf{u}^i \|_2^2 + \varphi(p)\}
\end{equation}
where $p$ is determined by the minimizer function $\varphi(.)$ with respect to $\phi(.)$.
\end{lemma}

Let $\mathbf{P}_{(v)} = \mathbf{F} - \mathbf{D}_{(v)}^{-1} \mathbf{A}_{(v)} \mathbf{F}$. According to the analysis for the $\ell_{2,1}$ norm in \cite{he20122}, if we define $\phi (x) = \sqrt{x^2 + \epsilon}$, we can replace $\| \mathbf{P}_{(v)} \|_{2,1} $ with $\sum_{j=1}^{n} \phi( \| \mathbf{p}_{(v)}^j \|_2 )$. Thus, based on Lemma \ref{lemma_L21norm}, we reformulate the objective function of Eq.~(\ref{eq: mvge}) as follows:
\begin{align}
\min_{\mathbf{F}} &~ \sum_{v=1}^{m} \alpha_{(v)} \text{Tr} \left (\mathbf{P}_{(v)}^{\mathrm{T}}\mathbf{Q}_{(v)} \mathbf{P}_{(v)}\right) \nonumber \\
\text{s.t.} & ~~\mathbf{F}^{\mathrm{T}}\mathbf{F}=\mathbf{I}_k
\label{eq:reform_obj}
\end{align}
where $\mathbf{Q}_{(v)} = Diag(\mathbf{q}_{(v)})$, and $\mathbf{q}_{(v)}$ is an auxiliary vector of the $\ell_{2,1}$ norm. The elements of $\mathbf{q}_{(v)}$ are computed by 
$ {q_{(v)}}_j = \frac{1}{2\sqrt{\| \mathbf{p}_{(v)}^j \|_2^2 + \epsilon}}$, where $\epsilon$ is a smoothing term and is usually set to be a small constant value (we set $\epsilon=10^{-4}$ in this paper).

Plugging $\mathbf{P}_{(v)} = \mathbf{F} - \mathbf{D}_{(v)}^{-1} \mathbf{A}_{(v)} \mathbf{F}$ into Eq.~(\ref{eq:reform_obj}), we can have the full form of the objective function with respect to $\mathbf{F}$ as 
\begin{align}
\min_{\mathbf{F}} &~ \sum_{v=1}^{m} \alpha_{(v)} \text{Tr} \left (\mathbf{F}^{\mathrm{T}}\mathbf{L}_{(v)} \mathbf{F}\right) \nonumber \\
\text{s.t.} & ~~\mathbf{F}^{\mathrm{T}}\mathbf{F}=\mathbf{I}_k
\label{eq:full_obj}
\end{align}
\noindent where $\mathbf{L}_{(v)}= \left ( \mathbf{I}_n - \mathbf{D}_{(v)}^{-1} \mathbf{A}_{(v)} \right )^{\mathrm{T}} \mathbf{Q}_{(v)} \left ( \mathbf{I}_n - \mathbf{D}_{(v)}^{-1} \mathbf{A}_{(v)} \right )$.

\subsection{An Auto-weighted Framework: MVGE-HD} \label{sec_framework}
In the literature of multi-view graph learning, adding a weight parameter for each view tend to be a common way for balancing the influence of different views of the data, and the choice of the parameter values is usually crucial to the final performance\cite{cai2013heterogeneous,li2015large,xia2010multiview}. The optimal parameter value tends to change for different datasets. Therefore, it is critical to avoid this problem and make the multi-view graph embedding approach more general to be applied to different datasets. Inspired by the auto-weighted multiple graph learning strategy proposed in \cite{nie2016parameter}, we further derive our objective function and propose an auto-weighted framework called MVGE-HD as follows. 

Following Eq.~(\ref{eq:full_obj}), we assume there is no weight parameters explicitly defined for each view, and we take the following form for the general framework 
\begin{align}
\min_{\mathbf{F}} &~ \sum_{v=1}^{m} \sqrt{\text{Tr} \left (\mathbf{F}^{\mathrm{T}}\mathbf{L}_{(v)} \mathbf{F}\right)} \nonumber \\
\text{s.t.} & ~~\mathbf{F}^{\mathrm{T}}\mathbf{F}=\mathbf{I}_k
\label{eq:sqrt_obj}
\end{align}

The Lagrange function of Eq.~(\ref{eq:sqrt_obj}) can be written as
\begin{align}
\min_{\mathbf{F}} &~ \sum_{v=1}^{m} \sqrt{\text{Tr} \left (\mathbf{F}^{\mathrm{T}}\mathbf{L}_{(v)} \mathbf{F}\right)} + \mathcal{G}\left (\mathbf{\Lambda}, \mathbf{F}\right ) 
\label{eq:langa_obj}
\end{align}
where $\mathbf{\lambda}$ is the Lagrange multiplier, and $\mathcal{G}(\mathbf{\Lambda}, \mathbf{F})$ represents the Lagrange term derived from the constraint.

Then we take the derivative of Eq.~(\ref{eq:langa_obj}) with respect to $\mathbf{F}$ and set the derivative to be zero. We will have
\begin{align}
\min_{\mathbf{F}} &~ \sum_{v=1}^{m} {\alpha}_{(v)}\frac{{\partial}\text{Tr} \left (\mathbf{F}^{\mathrm{T}}\mathbf{L}_{(v)} \mathbf{F}\right)}{{\partial}\mathbf{F}} + \frac{{\partial}\mathcal{G}\left (\mathbf{\Lambda}, \mathbf{F}\right )}{{\partial}\mathbf{F}} = 0 
\label{eq:derivative_obj}
\end{align}
where 
\begin{align}
{\alpha}_{(v)}=\frac{1}{2\sqrt{\text{Tr} \left (\mathbf{F}^{\mathrm{T}}\mathbf{L}_{(v)} \mathbf{F}\right)}}
\label{eq:alpha}
\end{align}

We can easily find that Eq.~(\ref{eq:derivative_obj}) can be regarded as the solution to the problem in Eq.~(\ref{eq:full_obj}) if ${\alpha}_{(v)}$ is set with a stationary value. However, as shown in Eq.~(\ref{eq:alpha}), the value of ${\alpha}_{(v)}$ depends on the variable $\mathbf{F}$. To solve this problem, we employ the alternating optimization scheme to update $\mathbf{F}$ and ${\alpha}_{(v)}$ alternately in an iterative manner. Given an initialized $\mathbf{F}$, we can compute the value for ${\alpha}_{(v)}$, according to Eq.~(\ref{eq:alpha}). Then the new ${\alpha}_{(v)}$ will be used consecutively to update $\mathbf{F}$ by solving Eq.~(\ref{eq:full_obj}), so on and so forth until convergence. After this iterative optimization process, we will obtain both the learned weight ${\alpha}_{(v)}$ and the multi-view graph embedding $\mathbf{F}$ for Eq.~(\ref{eq:full_obj}), which is the real problem we aim to solve.

In the above multi-view graph embedding problem, if view $v$ can provide much useful information, we say it is a good view, and the value of $\text{Tr}(\mathbf{F}^{\mathrm{T}}\mathbf{L}_{(v)} \mathbf{F})$ should be small. Based on Eq.~(\ref{eq:alpha}), the weight ${\alpha}_{(v)}$ will be large. Accordingly, a bad view will have a small weight. This indicates that the optimization scheme of the weights in our framework is reasonable.

Based on the above analysis, we can find that the proposed MVGE-HD framework can learn the multi-view graph embedding with hubs and the weight of each view simultaneously, thus can serve as a general framework for learning multi-view graph embedding on various datasets. The details of the optimization process and the convergence analysis of the framework will be introduced later in Section~\ref{sec_opt}.

%% file: optimize_04.tex
\section{Optimization}\label{sec_opt}
Following the analysis in Section~\ref{sec_method}, we present the iterative optimization process of MVGE-HD in this section. 
We start with the initialization of weight factor ${\alpha}_{(v)}$ for each view $v$, and set them to be $\frac{1}{m}$ equally. Now we compute $\mathbf{F}$ by solving the minimization problem~(\ref{eq:full_obj}). If we treat the $\sum_{v=1}^{m} \alpha_{(v)}\mathbf{L}_{(v)}$ in Eq.~(\ref{eq:full_obj}) as a Laplacian matrix $\widetilde{\mathbf{L}}$, based on the spectral analysis in \cite{von2007tutorial}, the optimal $\mathbf{F}$ can be computed by solving the eigenvector problem of the matrix
\begin{align}
\widetilde{\mathbf{L}}= \sum_{v=1}^{m} \alpha_{(v)} \left( \mathbf{I}_n - \mathbf{D}_{(v)}^{-1} \mathbf{A}_{(v)} \right )^{\mathrm{T}} \mathbf{Q}_{(v)} \left( \mathbf{I}_n - \mathbf{D}_{(v)}^{-1} \mathbf{A}_{(v)} \right )
\label{eq:L} 
\end{align}
Note that according to the illustration in Section~\ref{sec_method_A}, the diagonal matrix $\mathbf{Q}_{(v)}$ is dependent on $\mathbf{F}$. Therefore we need to compute $\mathbf{Q}_{(v)}$ first following its definition in Section \ref{sec_method_A} before updating $\mathbf{F}$. After we obtain the updated $\mathbf{F}$, we can use it to compute the weight factor ${\alpha}_{(v)}$ by Eq.~(\ref{eq:alpha}) for the next iteration, which will be used to compute $\mathbf{F}$ again following the same process discussed above. We summarize the overall optimization algorithm in Algorithm 1. 
\renewcommand{\algorithmicrequire}{\textbf{Input:}}
\renewcommand{\algorithmicensure}{\textbf{Output:}}
\begin{algorithm}[t]
\caption{MVGE-HD}
\label{alg:MVGE}
\begin{algorithmic}[1] 
\Require Affinity matrices for $m$ views $A =\{\mathbf{A}_{(1)},\mathbf{A}_{(2)},\cdots ,\mathbf{A}_{(m)}\}$; the dimension of the graph embedding $k$
\Ensure The graph embedding matrix $\mathbf{F}$, \\
    Initialize $\mathbf{F}_0$ s.t. $\mathbf{F}_0^{\mathrm{T}} \mathbf{F}_0 = \mathbf{I}_k, t \leftarrow 0$;
\While {not converge}
\State Compute ${\alpha}_{(v)_t}$ for $v = 1,\cdots, m$ by Eq.~(\ref{eq:alpha});
\State Set ${\mathbf{Q}_{(v)}}_t \leftarrow Diag(\frac{1}{2\sqrt{\| \mathbf{p}_{(v)}^j \|_2^2 + \epsilon}}) $;
\State Compute ${\mathbf{F}}_{t+1} $ by calculating the eigenvectors corresponding to the 2nd to $(k + 1)$-th smallest eigenvalues of matrix $\widetilde{\mathbf{L}}$ in Eq.~(\ref{eq:L});
\State $t \leftarrow t+1$;
\EndWhile
\end{algorithmic}
\end{algorithm}

Based on the analysis in \cite{nie2016parameter}, it is obvious that the solution in Algorithm~\ref{alg:MVGE} will converge to a local optimum of the problem ~(\ref{eq:sqrt_obj}), as the updated $\mathbf{F}$ in each iteration of Algorithm~\ref{alg:MVGE} monotonically decrease the objective function in Eq.~(\ref{eq:sqrt_obj}). For details about the theorem and proof, users can refer to the illustrations in \cite{nie2016parameter}.

%% file: experiment_05.tex
\section{Experiments and Analysis}\label{sec:exp}
\subsection{Data Collection and Preprocessing}
In this work, we use two real datasets as follows:
\begin{itemize}
\item \textit{Human Immunodeficiency Virus Infection (HIV)}: This dataset is collected from the Chicago Early HIV Infection Study at Northwestern University\cite{ragin2012structural}. This clinical study involves 77 subjects, 56 of which are early HIV patients (positive)  and the other 21 subjects are seronegative controls (negative). These two groups of subjects do not differ in demographic characteristics such as age, gender, racial composition and education level. This dataset contains both the functional magnetic resonance imaging (fMRI) and diffusion tensor imaging (DTI) for each subject, from which we can construct the fMRI and DTI brain networks. Then we can treat them as graphs with two views.
\item \textit{Bipolar}: This dataset consists of the fMRI and DTI image data of 52 bipolar I subjects who are in euthymia and 45 healthy controls with matched age and gender. The resting-state fMRI scan was acquired on a 3T Siemens Trio scanner using a T2*-weighted echo planar imaging (EPI) gradient-echo pulse sequence with integrated parallel acquisition technique (IPAT), set with TR = 2 sec, TE = 25 msec, flip angle = 78, matrix = 64x64, FOV = 192 mm, in-plane voxel size = 3x3 mm, slice thickness = 3 mm, 0.75 mm gap, and 30 total interleaved slices. Two TRs at the beginning of the scan were discarded to allow for scanner equilibration. There are 208 volumes acquired for the total sequence time of 7 min and 2 sec. Diffusion weighted MRI data were acquired on a Siemens 3T Trio scanner. 60 contiguous axial brain slices were collected using the following parameters: 64 diffusion-weighted (b = 1000s/\(mm^2\)) and 1 non-diffusion weighted scan; field of view (FOV) 190x190 mm; voxel size 2x2x2 mm; TR = 8400 ms; TE = 93 ms. In addition, high-resolution structural images were acquired using T1-weighted magnetization-prepared rapid-acquisition gradient echo (MPRAGE; FOV 250x250 mm; voxel size: 1x1x1 mm; TR = 1900 ms, TE = 2.26 ms, flip angle = 9 \textdegree).
\end{itemize}

We perform preprocessing on the HIV dataset using the standard process as illustrated in \cite{cao2015identifying}. First, we use the DPARSF toolbox\footnote{http://rfmri.org/DPARSF.} to process the fMRI data. We realign the images to the first volume, do the slice timing correction and normalization, and then use an 8-mm Gaussian kernel to smooth the image spatially. The band-pass filtering ($0.01$-$0.08$ Hz) and linear trend removing of the time series are also performed. We focus on the 116 anatomical volumes of interest (AVOI), each of which represents a specific brain region, and extract a sequence of responds from them. Finally, we construct a brain network with the 90 cerebral regions. Each node in the graph represents a brain region, and links are created based on the correlations between different brain regions. For the DTI data, we use FSL toolbox\footnote{http://fsl.fmrib.ox.ac.uk/fsl/fslwiki.} for the preprocessing and then construct the brain networks. The preprocessing includes distortion correction, noise filtering, repetitive sampling from the distributions of principal diffusion directions for each voxel. We parcellate the DTI images into the 90 regions same with fMRI via the propagation of the Automated Anatomical Labeling (AAL) on each DTI image \cite{tzourio2002automated}.

For the Bipolar dataset, the brain networks were constructed using the CONN\footnote{http://www.nitrc.org/projects/conn} toolbox \cite{whitfield2012conn}. The raw EPI images were first realigned and co-registered, after which we perform the normalization and smoothing. Then the confound effects from motion artifact, white matter, and CSF were regressed out of the signal. Finally, the brain networks were derived using the pairwise signal correlations based on the 82 labeled Freesurfer-generated cortical/subcortical gray matter regions.

\subsection{Baselines and Evaluation Metrics}
In brain network study, an important task is to use the graph connectivity features for neurological disorder analysis. As introduced above, both the HIV dataset and Bipolar dataset have the two-view brain networks of a group of subjects with neurological disorder and a group of normal controls. In this paper, to evaluate the effectiveness of the proposed MVGE-HD framework for brain network analysis, we apply MVGE-HD on each of the multi-view brain network instances in HIV dataset and Bipolar dataset, and then we use the learned multi-view graph embedding as the feature of each instance and use it for clustering the subjects in HIV dataset and Bipolar dataset, respectively. Then we evaluate the MVGE-HD approach by investigating how well the resulting multi-view graph embedding of MVGE-HD can help in separating the neurological disordered subjects and normal controls. In addition, we also look into the hubs learned by our framework and analyze them in the perspective of neuroscience.

We compared our MVGE-HD framework with seven other baseline methods on the HIV and Bipolar datasets. As our proposed framework is the first work on jointly learning multi-view graph embedding and hubs, there is no other existing method proposed for the same problem. Therefore, for the evaluation, we apply several state-of-the-art methods of multi-view graph embedding as baselines and adapt them for the problem here. 
\begin{itemize}
    \item \textbf{SingleBest} applies the single-view version of the proposed MVGE-HD framework (\ie, Eq.(\ref{eq: single_mvge})) on each single view and reports the best performance among them. 
    \item \textbf{SEC} is a single view spectral embedding clustering approach proposed in \cite{nie2011spectral}. It imposes a linearity regularization on the spectral clustering model and uses both local and global discriminative information for the embedding.
    \item \textbf{CoRegSc} is the co-regularized based multi-view spectral clustering framework proposed in \cite{kumar2011co}. The centroid based approach is applied for the multi-view graph embedding task here. 
    \item \textbf{MMSC} is the multi-modal spectral clustering method proposed by \cite{cai2011heterogeneous}. It aims to learn a commonly shared graph Laplacian matrix by unifying different views.
    \item \textbf{AMGL} is a recently proposed multi-view spectral learning approach~\cite{nie2016parameter} that can automatically learn an optimal weight for each graph without introducing additive parameters. 
    \item \textbf{BC$\textbf{+}$CoRegSc} is the method we combined with Betweenness Centrality~\cite{brandes2001faster} and CoRegSc for evaluating if the hubs detected would help improve the multi-view graph embedding of CoRegSc, and also for comparing with our method. Betweenness Centrality (BC) is a popular method for hub detection in both social network and brain network. We first apply BC on each view of the data to obtain the top-$k$ hubs, and then we remove their connections with other nodes in the graph by setting the corresponding values in affinity matrix to be 0. Then we run CoRegSc with the new affinity matrices from all the views for learning the multi-view graph embedding. 
    \item \textbf{MVGE-HD}{\textbf{*}} represents the proposed approach in Eq.~(\ref{eq: mvge}) without auto-weighted ability. We set the weight parameter $\alpha_{(v)}$ as $0.5$ for each of the two views, and evaluate the performance for the comparison with the auto-weighted version of the proposed framework.
    \item \textbf{MVGE-HD} is the proposed auto-weighted framework for multi-view graph embedding with hub detection.  
\end{itemize}

After we run each of the above algorithms on the data, we will obtain a multi-view graph embedding matrix $\mathbf{F}$ for each multi-view brain network instance. To facilitate the clustering of the instances, we use the following equation to compute the similarity between each pair of instances~\cite{frey2007clustering}. 
\begin{align}
    {s_{ij}} = - \sqrt{\text{Tr} \left( \left(\mathbf{F}_i - \mathbf{F}_j \right)^{\mathrm{T}} \left(\mathbf{F}_i - \mathbf{F}_j \right) \right)}
\end{align}
where $\mathbf{F}_i$ is the multiv-view graph embedding of instance $i$ and $\mathbf{F}_j$ is the multi-view graph embedding of instance $j$. 

Then we apply the standard spectral clustering procedure \cite{shi2000normalized} for the clustering of the brain network instances. For the k-means clustering step in the experiment, we use the Litekmeans \cite{cai2011litekmeans} implementation. 

As the weight factor $\alpha_{(v)}$ in the proposed MVGE-HD framework is auto-tuned, for fair comparisons of the baseline methods, we tune parameters for each of the baseline methods, and report their performance with the optimal parameter settings. The optimal value for the multi-view graph embedding dimension $k$ is selected by the grid search from $\{5, 6, \cdots, 15\}$. For each experiment, we repeat 20 times and report the mean value with the standard deviation (std) as the results. In the clustering stage of the brain network instances, we set the number of clusters to be 2, as there are two possible labels (\ie, patient and normal control) in the HIV and Bipolar datasets. 

We adopt the following measures for the evaluation. 

\begin{itemize}
\item {\emph{Accuracy (ACC)}}. Let $c_i$ represent the clustering label result of the clustering algorithm and $y_i$ represent the ground truth label of the two-view brain network instance $i$. Then \emph{Accuracy} is defined as:
\begin{align}
    Accuracy = \frac{\sum_{i=1}^{n} \delta(y_i, map(c_i))}{n}
\end{align}
where $\delta$ is the Kronecker delta function, and $map(c_i)$ is the best mapping function that permutes clustering labels to match the ground truth labels using the KuhnMunkres algorithm~\cite{kuhn1955hungarian}. A larger \emph{ACC} indicates better clustering performance.
\item{\emph{Normalized Mutual Information (NMI)}}. Normalized Mutual Information is a measure used to evaluate the mutual information entropy between the resulted cluster labels and the ground truth labels. For any two variables $X$ and $Y$, \textit{NMI} is defined as: 
\begin{align}
   \textit{NMI} = \frac{I(X,Y)}{\sqrt{H(X)H(Y)}} 
\end{align}
where $I(X,Y)$ computes the mutual information between $X$ and $Y$, and $H(X)$ and $H(Y)$ are the entropies of $X$ and $Y$, respectively. The larger the \textit{NMI} value, the better the clustering performance.
\end{itemize}

\begin{table}[t]
\caption{Results on HIV dataset (mean $\pm$ std).}
\label{tab:hiv}
\centering
\begin{tabular}{lcc}
\toprule
Methods        &\emph{ACC}   &\emph{NMI}\\
\midrule
SingleBest              
&$0.579 \pm 0.011$      &$0.086 \pm 0.009$ \\
SEC              
&$0.552 \pm 0.010$      &$0.058 \pm 0.011$ \\
AMGL                     
&$0.582 \pm 0.002$      &$0.091 \pm 0.006$ \\
MMSC                  
&$0.586 \pm 0.013$      &$0.105 \pm 0.010$ \\
CoRegSc
&$0.625 \pm 0.012$      &$0.163 \pm 0.015$ \\
BC\text{+}CoRegSc
&$0.635 \pm 0.009$      &$0.190 \pm 0.008$ \\
MVGE-HD*
&$0.613 \pm 0.010$      &$0.152 \pm 0.008$ \\
MVGE-HD
&\textbf{0.701} $\pm$ \textbf{0.012}  &\textbf{0.261} $\pm$ \textbf{0.011} \\
\bottomrule
\end{tabular}
\end{table}

\begin{table}[t]
\caption{Results on Bipolar dataset (mean $\pm$ std).}
\label{tab:bipolar}
\centering
\begin{tabular}{lcc}
\toprule
Methods        &\emph{ACC}   &\emph{NMI}\\
\midrule
SingleBest              
&$0.565 \pm 0.012$      &$0.074 \pm 0.009$ \\
SEC              
&$0.549 \pm 0.012$      &$0.067 \pm 0.008$ \\
AMGL                     
&$0.563 \pm 0.001$      &$0.088 \pm 0.006$ \\
MMSC                  
&$0.608 \pm 0.014$      &$0.119 \pm 0.011$ \\
CoRegSc
&$0.637 \pm 0.011$      &$0.194 \pm 0.013$ \\
BC\text{+}CoRegSc
&$0.641 \pm 0.012$      &$0.203 \pm 0.009$ \\
MVGE-HD*
&$0.628 \pm 0.010$      &$0.175 \pm 0.009$ \\
MVGE-HD
&\textbf{0.712} $\pm$ \textbf{0.010}  &\textbf{0.266} $\pm$ \textbf{0.011} \\
\bottomrule
\end{tabular}
\end{table}

\subsection{Performance Analysis}
Table~\ref{tab:hiv} and Table~\ref{tab:bipolar} show the the clustering performance by using the multi-view graph embedding obtained with each of the seven methods on the HIV dataset and Bipolar dataset, respectively. As we can see from Table~\ref{tab:hiv} and Table~\ref{tab:bipolar}, the multi-view graph embedding obtained by the proposed MVGE-HD framework results in the best clustering performance on both of the two datasets in terms of \textit{accuracy} and \textit{NMI}. Among the eight methods, the SingleBest and SEC are the only two single-view graph embedding methods, and we can find that they both achieve lower accuracy compared with most of the multi-view methods, although the SingleBest performs slightly better than AMGL on Bipolar dataset in terms of \textit{accuracy}. This indicates that the information combined from multiple views can lead to a better graph embedding result than that of a single view. Comparing with SEC, the SingleBest method achieves higher \textit{accuracy} and \textit{NMI} on both datasets. This is probably because that the SingleBest considers the hubs when doing graph embedding, while SEC only focuses on the spectral analysis for the embedding. In the experiment, the best performance of SingleBest and SEC both occur in the fMRI brain networks, which means that fMRI data provide more discriminative information for SingleBest and SEC.

Among the six multi-view graph embedding methods, the BC\textit{+}CoRegSc, MVGE-HD* and the MVGE-HD consider the hubs when performing the multi-view graph embedding, while the three other methods do not. We can see that all the three methods that consider hub detection achieve better performance than the other three methods. This implies that detecting the hubs and reducing their effect in the multi-view graph embedding process benefit the task, and the multi-view graph embedding obtained in this case tend to be more discriminative for the analysis of multiple graph instances. Meanwhile, we can see that, although the BC\textit{+}CoRegSc method performs better than the other baselines, the accuracy it achieves is still much lower than that of our proposed MVGE-HD approach. This is mainly due to the fact that the hub detection stage and multi-view graph embedding stage is separately done by BC\textit{+}CoRegSc. The hubs detected by BC may not correspond to the hubs implied by the multi-view graph embedding derived from CoRegSc, although some of the hubs detected may be helpful for the multi-view graph embedding stage by CoRegSc. Comparatively, in the proposed MVGE-HD framework, the hub detection is done along with the multi-view graph embedding, and by shrinking the embedding row vector of the potential hubs to zero, the resulted multi-view graph embedding would reflect a more discriminative node clustering structure of the graph. In addition, we find that the MVGE-HD* method, which is the version of MVGE-HD with an equal weight factor as $0.5$ for each view, achieves much lower \textit{accuracy} and \textit{NMI} compared to the auto-weighted MVGE-HD framework. This indicates that the auto-weighted ability is very important for the multi-view graph embedding with hub detection task. In the multi-view learning process, different views may exert different levels of influence on the multi-view task, and the optimal weight for each view often varies from dataset to dataset. Therefore, the auto-weighted ability of the proposed MVGE-HD framework enables it be easily applied for different datasets.

\begin{figure}[t]
\centering
    \begin{subfigure}[\emph{HIV}]{		
            \includegraphics[width=.45\linewidth] {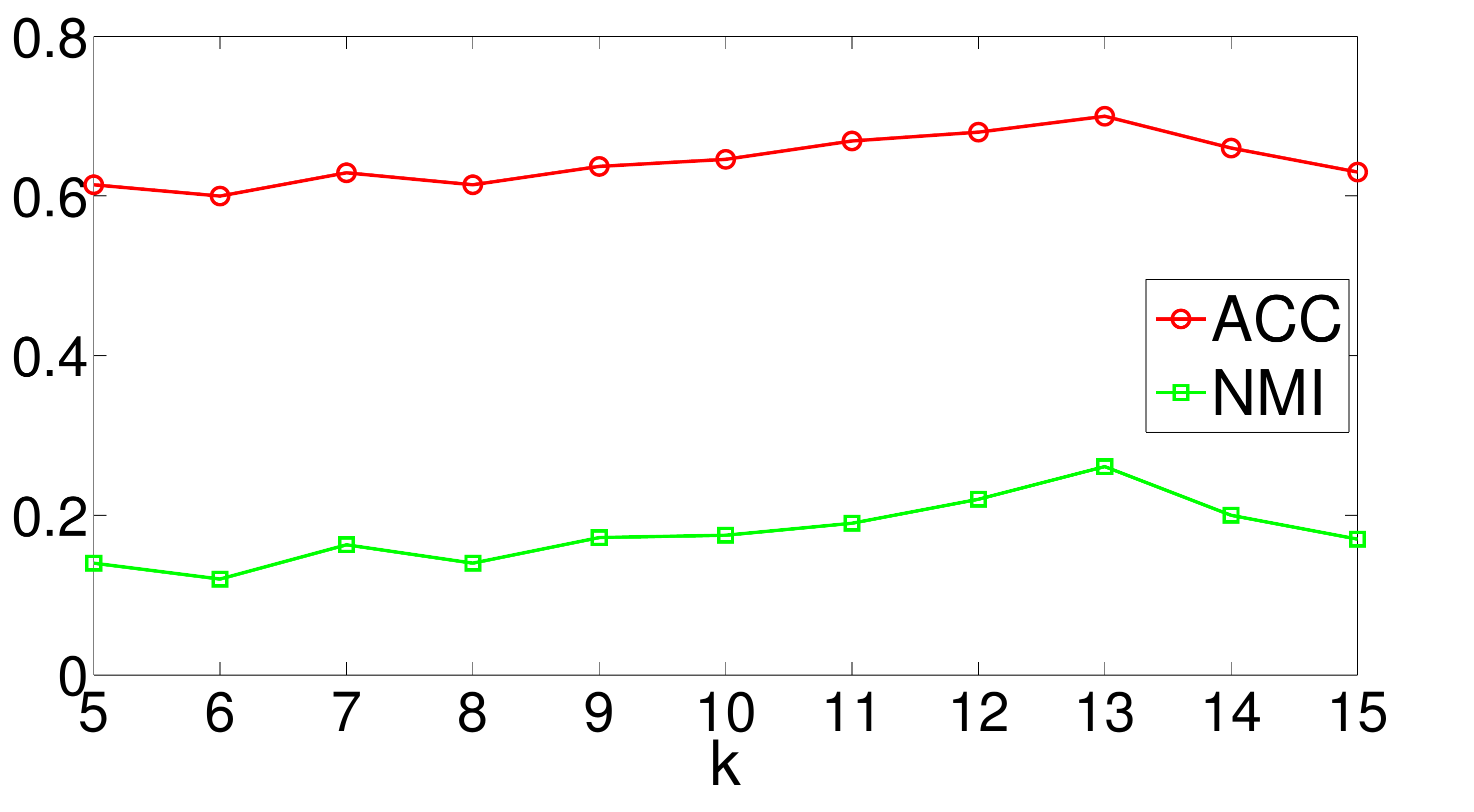}
		    \label{fig:para_hiv}
		}%
		\end{subfigure}
		\begin{subfigure}[\emph{Bipolar}]{
		\includegraphics[width=.45\linewidth]{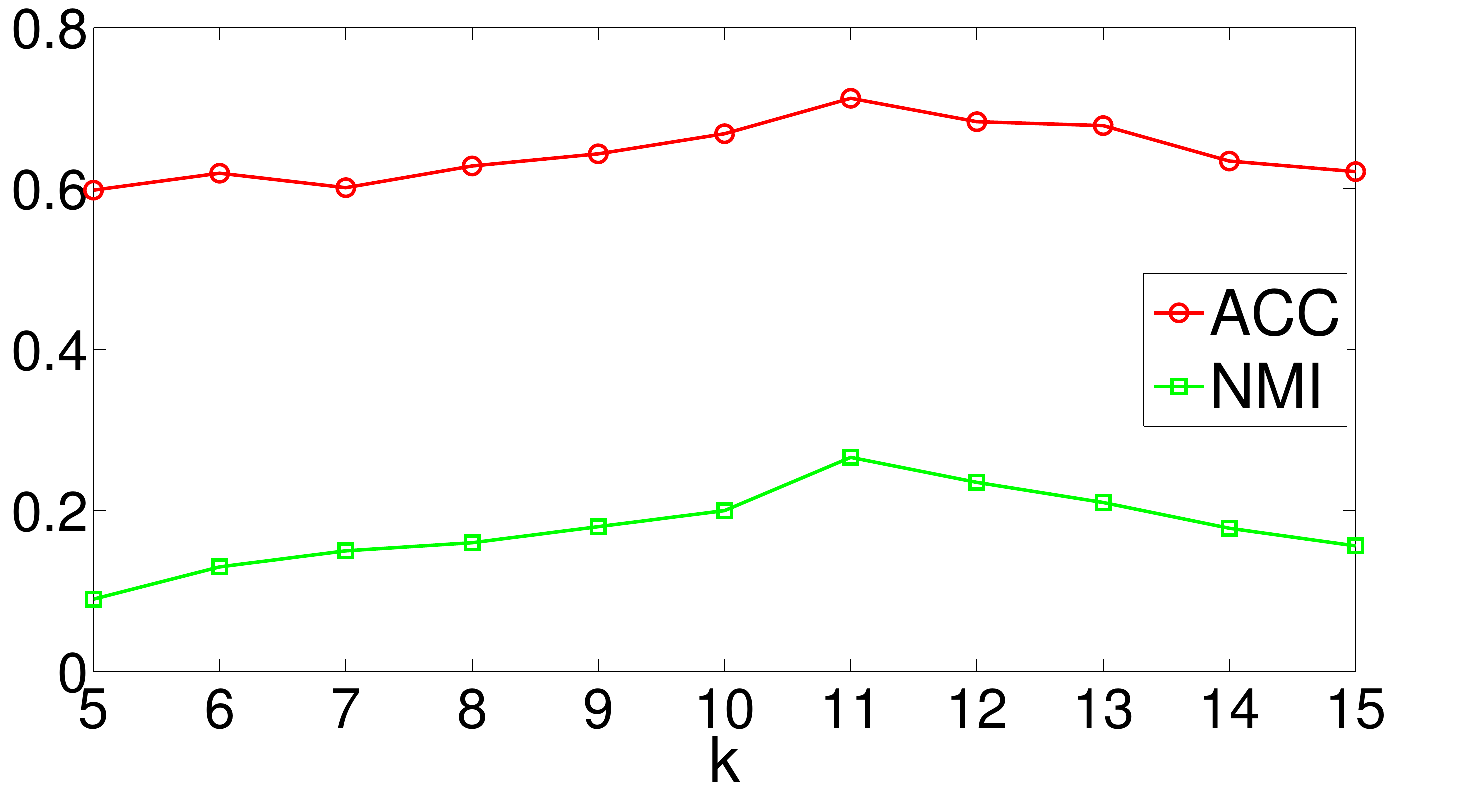}
		\label{fig:para_bipolar}
		}%
		\end{subfigure}  
  \caption{\emph{Accuracy} and \emph{NMI} with different $c$}
  \label{fig:acc_nmi}
\end{figure}

In the proposed MVGE-HD framework, the only parameter  is the dimension of multi-view graph embedding, which is the $k$ introduced earlier. Now we evaluate the sensitivity of MVGE-HD to different values of $k$. Fig.~\ref{fig:para_hiv} and Fig.~\ref{fig:para_bipolar}  show the performance of MVGE-HD corresponding to the $k$ values ranging from $5$ to $15$ on the HIV dataset and Bipolar dataset, respectively. As we can see from the figures, the value of $k$ affects the performance in both \textit{accuracy} and \textit{NMI}. For the HIV dataset, the best \textit{accuracy} and \textit{NMI} are achieved when $k = 13$, while the best \textit{accuracy} and \textit{NMI} occurs at $k = 11$ for the Bipolar dataset. The changing of \textit{accuracy} and \textit{NMI} with respect to different $k$ values have similar trend on both of the two datasets. With the increase of $k$ value, the performance first keeps rising up until it reaches the peak, and then it starts to decline. This indicates that when the dimension of multi-view graph embdding is too low, it could not capture enough structure information of the graph, leading to poor performance. When the dimension is set to be a large value, it may contain much redundant information, thus being less discriminative to be used for the clustering task. Therefore, the dimension of the multi-view graph embedding for the MVGE-HD framework should be set based on the application scenarios. 

\begin{figure}[t]
\centering
        \begin{subfigure}[normal control]{\includegraphics[width=.45\linewidth] {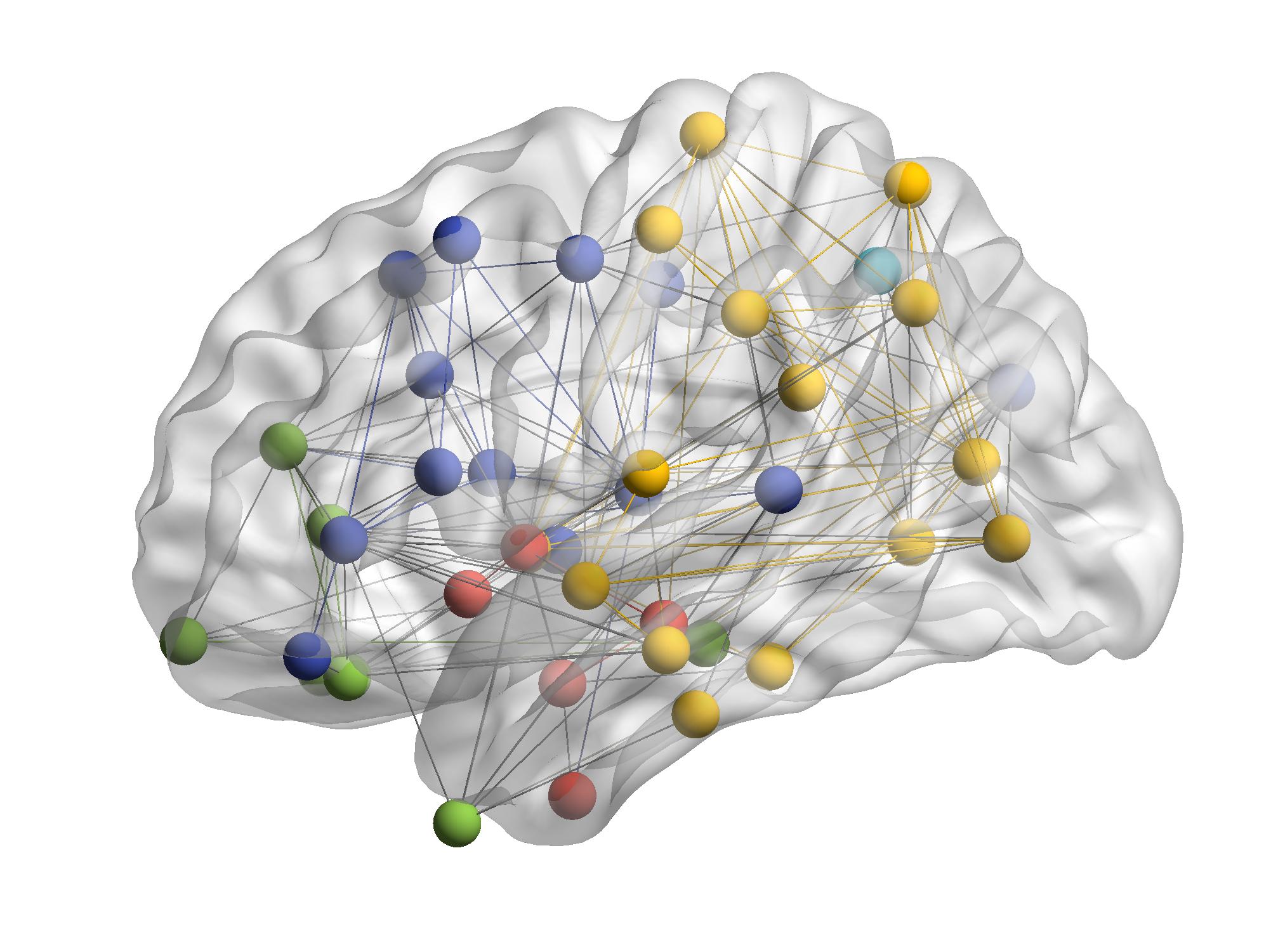}
		\label{fig:bpN}
		}%
		\end{subfigure}
		\begin{subfigure}[bipolar subject]{
		\includegraphics[width=.45\linewidth]{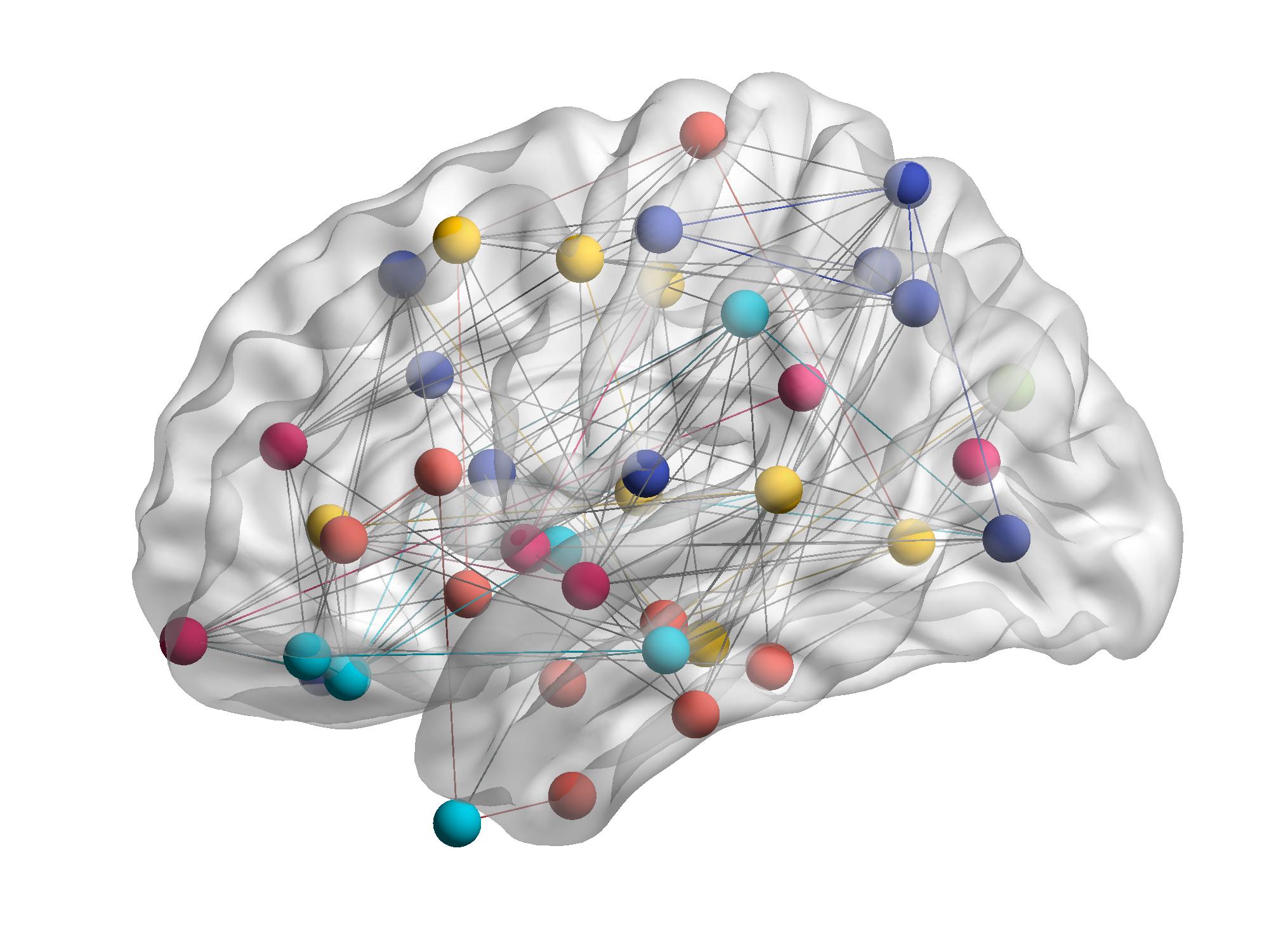}
		\label{fig:bpP}
		}%
		\end{subfigure}
   \vspace{0.5em}
		\caption{Comparison of the brain region clusters resulted from MVGE-HD on the brain networks of a normal control and a bipolar subject}
		\centering
		\label{fig:compare_bipolar}
\end{figure}

To evaluate the effectiveness of the proposed MVGE-HD framework for brain region clustering analysis, after we obtain the multi-view graph embedding of all the brain networks, we further apply the k-means algorithm with k equal to the dimension value $k$ on the row vectors of the multi-view graph embedding for each brain network instance and then we visualize the clustering results using the Brain Net Viewer toolbox \cite{xia2013brainnet}. Fig.~\ref{fig:compare_bipolar} shows an example of the resulted visualized brain network with 6 clusters (\ie, $k = 6$) of a normal control and a bipolar subject. In the figures, each node represents a brain region in the network, and the nodes with the same color refer to the brain regions that have been clustered into the same group, and the edges represent the connections between different brain regions.

As we can see from Fig.~\ref{fig:bpN}, the clusters in the brain network of the normal control look quite clear, while the clusters in brain network of the bipolar subject as shown in Fig.~\ref{fig:bpP} is very messy. This indicates that the collaborations of different brain regions are well-organized for the normal control, as the regions close to each other in the brain are usually highly correlated and tend to collaborate more in brain activities. However, for the bipolar subject, the collaborations of the brain regions are probably in some kind of disorder, leading to the messy clusters as shown in Fig.~\ref{fig:bpP}. Moreover, the big difference between the clustering maps of the two networks is probably partially due to the difference of their hubs. Since MVGE-HD can detect the multi-view hubs and adjust the multi-view embedding with the hubs, when the hubs of the neurological disordered brain networks are different from those of normal people, the multi-view graph embedding guided by the hub detection of MVGE-HD would also be different for them. These observations coincide well with the findings about hubs in neuroscience study \cite{crossley2014hubs}. In addition, from Fig.~\ref{fig:bpN}, we can also find that although some boundary nodes between the clusters have quite a few cross edges, which means they are the hubs in the brain network, the clusters resulted from MVGE-HD are not blurred by these nodes. This implies that our MVGE-HD approach can reduce the influence of these hubs, thus leading to clear cluster boundaries and discriminative clustering structure for the brain networks.

%% file: relatedwork_06.tex
\section{Related Work}\label{sec:related}
Our work relates to several branches of studies, which include multi-view graph learning, hub detection and brain network analysis. 

\subsection{Multi-View Graph Embedding}
Multi-view graph embedding has been a widely studied topic for the multi-view learning community in recent years. The key issue in multi-view graph embedding is how to combine the multiple views, so that both the consensus and complementary information across different views can be utilized for learning the embedding. The existing methods in this field can be divided into three categories. In the first category, the multiple views are often combined via integrating the affinity matrix or other graph features of each view. For example, in \cite{huang2012affinity}, a multi-view spectral clustering algorithm is proposed based on affinity aggregation, which seeks for an optimal combination of affinity matrices for the spectral clustering across multiple views. In \cite{li2015large}, a large-scale multi-view spectral clustering method is introduced, which uses local manifold fusion to integrate heterogeneous features of graphs. The second category of works aim to learn a new Laplacian matrix by combining the Laplacian matrices of different views. For instance, a multi-modal spectral clustering algorithm is presented in \cite{cai2011heterogeneous} to learn a commonly shared graph Laplacian matrix by unifying different views. For the works in the third category, they aim to obtain a consistent clustering across all the views by adjusting the clustering along with learning features from the multiple views. In \cite{papalexakis2013more}, two solutions of multi-view graph embedding are proposed, which use the minimum description length and tensor decomposition principles respectively for graph clustering across multiple views. Another classic method for finding a consistent clustering across the multiple views is the co-regularized multi-view spectral clustering method proposed in \cite{kumar2011co}, which is also a baseline method we use in the experiments.

\subsection{Hub Detection}
Hub detection is also an essential research topic in graph mining. In the past decade, quite a few of works have been done in this area. Some of them focus on the structural hole detection problem for social network analysis \cite{he2016joint, rezvani2015identifying}. In \cite{rezvani2015identifying}, they propose a method based on bounded inverse closeness centrality for analyzing the structural hole spanners, which are viewed as the vertices that can result in the maximum increase on the mean distance of the network if they are removed. In \cite{he2016joint}, a model called HAM is proposed for jointly detecting the communities and structural holes in social networks. They show that by removing the detected structural hole spanners, the quality of the learned communities can be improved. Some other works aim to use the hub detection measures for neuroscience study. For example, a review of network hubs in human brain is presented in \cite{van2013network}, and the rich-club organization of the human connectome is studied in \cite{van2011rich}, which illustrate the important role that hubs play in human brains. 

\subsection{Brain Network Analysis}
Brain network analysis is a prominent emphasis area in the field of medical data mining. So far, the researchers in this field aim to study the connectivity of neural systems at different levels involving both global and local structure information of the connections \cite{kaiser2011tutorial}. Brain network analysis has been the focus of intense investigation owing to the tremendous potential to provide more comprehensive understanding of normal brain function and to yield new insights concerning many different brain disorders \cite{sporns2005human,cao2017tbne,ma2016spatio}. Most connectome analyses, however, aim to learn the structure from brain networks based on an individual neuroimaging modality \cite{cao2015identification,kuo2015unified}. For example, in \cite{cao2015identification}, the identification of discriminative subgraph patterns is studied on fMRI brain networks for bipolar affective disorder analysis. In \cite{ma2016multi}, a multi-graph clustering method is proposed based on interior-node clustering for connectome analysis in fMRI resting-state networks. Although some recent work \cite{cahill2016multiple} use multi-view brain networks in connectome analysis, they focus on the group-wise functional community detection problem instead of doing multi-view graph embedding of each subject. Here, we apply the proposed multi-view graph embedding on each subject, which further facilitates the clustering of all the subjects, thus providing a more comprehensive strategy for further neurological disorder identification.

%% file: conclusion_07.tex
\section{Conclusion} \label{sec_conclusion}
In this paper, we present MVGE-HD, an auto-weighted framework of Multi-view Graph Embedding with Hub Detection for brain network analysis. We incorporate the hub detection task into the multi-view graph embedding framework so that the two tasks could benefit each other. The MVGE-HD framework learns a unified graph embedding across all the views while reducing the potential influence of the hubs on blurring the boundaries between node clusters in the graph, thus leading to a clear and discriminative node clustering structure for the graph. The extensive experimental results on two real multi-view brain network datasets (\ie, HIV and Bipolar) demonstrate the effectiveness and the superior performance of the proposed framework for brain network analysis. 

%% file: main.bbl
\begin{thebibliography}{10}
\providecommand{\url}[1]{#1}
\csname url@samestyle\endcsname
\providecommand{\newblock}{\relax}
\providecommand{\bibinfo}[2]{#2}
\providecommand{\BIBentrySTDinterwordspacing}{\spaceskip=0pt\relax}
\providecommand{\BIBentryALTinterwordstretchfactor}{4}
\providecommand{\BIBentryALTinterwordspacing}{\spaceskip=\fontdimen2\font plus
\BIBentryALTinterwordstretchfactor\fontdimen3\font minus
  \fontdimen4\font\relax}
\providecommand{\BIBforeignlanguage}[2]{{%
\expandafter\ifx\csname l@#1\endcsname\relax
\typeout{** WARNING: IEEEtran.bst: No hyphenation pattern has been}%
\typeout{** loaded for the language `#1'. Using the pattern for}%
\typeout{** the default language instead.}%
\else
\language=\csname l@#1\endcsname
\fi
#2}}
\providecommand{\BIBdecl}{\relax}
\BIBdecl

\bibitem{ma2017multi}
G.~Ma, L.~He, C.-T. Lu, W.~Shao, P.~S. Yu, A.~D. Leow, and A.~B. Ragin,
  ``Multi-view clustering with graph embedding for connectome analysis,'' in
  \emph{Proceedings of the 26th ACM International on Conference on Information
  and Knowledge Management}.\hskip 1em plus 0.5em minus 0.4em\relax ACM, 2017.

\bibitem{kumar2011co}
A.~Kumar, P.~Rai, and H.~Daume, ``Co-regularized multi-view spectral
  clustering,'' in \emph{Advances in neural information processing systems},
  2011, pp. 1413--1421.

\bibitem{papalexakis2013more}
E.~E. Papalexakis, L.~Akoglu, and D.~Ience, ``Do more views of a graph help?
  community detection and clustering in multi-graphs,'' in \emph{Information
  fusion (FUSION), 2013 16th international conference on}.\hskip 1em plus 0.5em
  minus 0.4em\relax IEEE, 2013, pp. 899--905.

\bibitem{cai2011heterogeneous}
X.~Cai, F.~Nie, H.~Huang, and F.~Kamangar, ``Heterogeneous image feature
  integration via multi-modal spectral clustering,'' in \emph{Computer Vision
  and Pattern Recognition (CVPR), 2011 IEEE Conference on}.\hskip 1em plus
  0.5em minus 0.4em\relax IEEE, 2011, pp. 1977--1984.

\bibitem{huang2012affinity}
H.-C. Huang, Y.-Y. Chuang, and C.-S. Chen, ``Affinity aggregation for spectral
  clustering,'' in \emph{Computer Vision and Pattern Recognition (CVPR), 2012
  IEEE Conference on}.\hskip 1em plus 0.5em minus 0.4em\relax IEEE, 2012, pp.
  773--780.

\bibitem{li2015large}
Y.~Li, F.~Nie, H.~Huang, and J.~Huang, ``Large-scale multi-view spectral
  clustering via bipartite graph.'' in \emph{AAAI}, 2015, pp. 2750--2756.

\bibitem{van2013network}
M.~P. van~den Heuvel and O.~Sporns, ``Network hubs in the human brain,''
  \emph{Trends in cognitive sciences}, vol.~17, no.~12, pp. 683--696, 2013.

\bibitem{he2016joint}
L.~He, C.-T. Lu, J.~Ma, J.~Cao, L.~Shen, and P.~S. Yu, ``Joint community and
  structural hole spanner detection via harmonic modularity,'' in
  \emph{Proceedings of the 22nd ACM SIGKDD International Conference on
  Knowledge Discovery and Data Mining}.\hskip 1em plus 0.5em minus 0.4em\relax
  ACM, 2016, pp. 875--884.

\bibitem{crossley2014hubs}
N.~A. Crossley, A.~Mechelli, J.~Scott, F.~Carletti, P.~T. Fox, P.~McGuire, and
  E.~T. Bullmore, ``The hubs of the human connectome are generally implicated
  in the anatomy of brain disorders,'' \emph{Brain}, vol. 137, no.~8, pp.
  2382--2395, 2014.

\bibitem{belkin2001laplacian}
M.~Belkin and P.~Niyogi, ``Laplacian eigenmaps and spectral techniques for
  embedding and clustering.'' in \emph{NIPS}, 2001.

\bibitem{fu2012graph}
Y.~Fu and Y.~Ma, \emph{Graph embedding for pattern analysis}.\hskip 1em plus
  0.5em minus 0.4em\relax Springer Science \& Business Media, 2012.

\bibitem{yan2007graph}
S.~Yan, D.~Xu, B.~Zhang, H.-J. Zhang, Q.~Yang, and S.~Lin, ``Graph embedding
  and extensions: a general framework for dimensionality reduction,''
  \emph{IEEE transactions on pattern analysis and machine intelligence},
  vol.~29, no.~1, pp. 40--51, 2007.

\bibitem{ma2016multi}
G.~Ma, L.~He, B.~Cao, J.~Zhang, P.~S. Yu, and A.~B. Ragin, ``Multi-graph
  clustering based on interior-node topology with applications to brain
  networks,'' in \emph{Joint European Conference on Machine Learning and
  Knowledge Discovery in Databases}.\hskip 1em plus 0.5em minus 0.4em\relax
  Springer, 2016, pp. 476--492.

\bibitem{he20122}
R.~He, T.~Tan, L.~Wang, and W.-S. Zheng, ``l 2, 1 regularized correntropy for
  robust feature selection,'' in \emph{Computer Vision and Pattern Recognition
  (CVPR), 2012 IEEE Conference on}.\hskip 1em plus 0.5em minus 0.4em\relax
  IEEE, 2012, pp. 2504--2511.

\bibitem{cai2013heterogeneous}
X.~Cai, F.~Nie, W.~Cai, and H.~Huang, ``Heterogeneous image features
  integration via multi-modal semi-supervised learning model,'' in
  \emph{Proceedings of the IEEE International Conference on Computer Vision},
  2013, pp. 1737--1744.

\bibitem{xia2010multiview}
T.~Xia, D.~Tao, T.~Mei, and Y.~Zhang, ``Multiview spectral embedding,''
  \emph{IEEE Transactions on Systems, Man, and Cybernetics, Part B
  (Cybernetics)}, vol.~40, no.~6, pp. 1438--1446, 2010.

\bibitem{nie2016parameter}
F.~Nie, J.~Li, X.~Li \emph{et~al.}, ``Parameter-free auto-weighted multiple
  graph learning: A framework for multiview clustering and semi-supervised
  classification.''\hskip 1em plus 0.5em minus 0.4em\relax International Joint
  Conferences on Artificial Intelligence, 2016.

\bibitem{von2007tutorial}
U.~Von~Luxburg, ``A tutorial on spectral clustering,'' \emph{Statistics and
  computing}, vol.~17, no.~4, pp. 395--416, 2007.

\bibitem{ragin2012structural}
A.~B. Ragin, H.~Du, R.~Ochs, Y.~Wu, C.~L. Sammet, A.~Shoukry, and L.~G.
  Epstein, ``Structural brain alterations can be detected early in hiv
  infection,'' \emph{Neurology}, vol.~79, no.~24, pp. 2328--2334, 2012.

\bibitem{cao2015identifying}
B.~Cao, X.~Kong, J.~Zhang, P.~S. Yu, and A.~B. Ragin, ``Identifying hiv-induced
  subgraph patterns in brain networks with side information,'' \emph{Brain
  informatics}, vol.~2, no.~4, pp. 211--223, 2015.

\bibitem{tzourio2002automated}
N.~Tzourio-Mazoyer, B.~Landeau, D.~Papathanassiou, F.~Crivello, O.~Etard,
  N.~Delcroix, B.~Mazoyer, and M.~Joliot, ``Automated anatomical labeling of
  activations in spm using a macroscopic anatomical parcellation of the mni mri
  single-subject brain,'' \emph{Neuroimage}, vol.~15, no.~1, pp. 273--289,
  2002.

\bibitem{whitfield2012conn}
S.~Whitfield-Gabrieli and A.~Nieto-Castanon, ``Conn: a functional connectivity
  toolbox for correlated and anticorrelated brain networks,'' \emph{Brain
  connectivity}, vol.~2, no.~3, pp. 125--141, 2012.

\bibitem{nie2011spectral}
F.~Nie, Z.~Zeng, I.~W. Tsang, D.~Xu, and C.~Zhang, ``Spectral embedded
  clustering: A framework for in-sample and out-of-sample spectral
  clustering,'' \emph{IEEE Trans on Neural Networks}, vol.~22, no.~11, pp.
  1796--1808, 2011.

\bibitem{brandes2001faster}
U.~Brandes, ``A faster algorithm for betweenness centrality,'' \emph{Journal of
  mathematical sociology}, vol.~25, no.~2, pp. 163--177, 2001.

\bibitem{frey2007clustering}
B.~J. Frey and D.~Dueck, ``Clustering by passing messages between data
  points,'' \emph{Science}, vol. 315, no. 5814, pp. 972--976, 2007.

\bibitem{shi2000normalized}
J.~Shi and J.~Malik, ``Normalized cuts and image segmentation,'' \emph{IEEE
  Transactions on pattern analysis and machine intelligence}, vol.~22, no.~8,
  pp. 888--905, 2000.

\bibitem{cai2011litekmeans}
D.~Cai, ``Litekmeans: the fastest matlab implementation of kmeans,''
  \emph{Software available at: http://www. zjucadcg.
  cn/dengcai/Data/Clustering. html}, 2011.

\bibitem{kuhn1955hungarian}
H.~W. Kuhn, ``The hungarian method for the assignment problem,'' \emph{Naval
  research logistics quarterly}, vol.~2, no. 1-2, pp. 83--97, 1955.

\bibitem{xia2013brainnet}
M.~Xia, J.~Wang, and Y.~He, ``Brainnet viewer: a network visualization tool for
  human brain connectomics,'' \emph{PloS one}, vol.~8, no.~7, p. e68910, 2013.

\bibitem{rezvani2015identifying}
M.~Rezvani, W.~Liang, W.~Xu, and C.~Liu, ``Identifying top-k structural hole
  spanners in large-scale social networks,'' in \emph{Proceedings of the 24th
  ACM International on Conference on Information and Knowledge
  Management}.\hskip 1em plus 0.5em minus 0.4em\relax ACM, 2015, pp. 263--272.

\bibitem{van2011rich}
M.~P. Van Den~Heuvel and O.~Sporns, ``Rich-club organization of the human
  connectome,'' \emph{Journal of Neuroscience}, vol.~31, no.~44, pp.
  15\,775--15\,786, 2011.

\bibitem{kaiser2011tutorial}
M.~Kaiser, ``A tutorial in connectome analysis: topological and spatial
  features of brain networks,'' \emph{Neuroimage}, vol.~57, no.~3, pp.
  892--907, 2011.

\bibitem{sporns2005human}
O.~Sporns, G.~Tononi, and R.~K{\"o}tter, ``The human connectome: a structural
  description of the human brain,'' \emph{PLoS Comput Biol}, vol.~1, no.~4, p.
  e42, 2005.

\bibitem{cao2017tbne}
B.~Cao, L.~He, X.~Wei, M.~Xing, P.~S. Yu, H.~Klumpp, and A.~D. Leow, ``t-{BNE}:
  Tensor-based brain network embedding,'' in \emph{Proceedings of SIAM
  International Conference on Data Mining (SDM)}, 2017.

\bibitem{ma2016spatio}
G.~Ma, L.~He, C.-T. Lu, P.~S. Yu, L.~Shen, and A.~B. Ragin, ``Spatio-temporal
  tensor analysis for whole-brain fmri classification,'' in \emph{Proceedings
  of the 2016 SIAM International Conference on Data Mining}.\hskip 1em plus
  0.5em minus 0.4em\relax SIAM, 2016, pp. 819--827.

\bibitem{cao2015identification}
B.~Cao, L.~Zhan, X.~Kong, P.~S. Yu, N.~Vizueta, L.~L. Altshuler, and A.~D.
  Leow, ``Identification of discriminative subgraph patterns in fmri brain
  networks in bipolar affective disorder,'' in \emph{International Conference
  on Brain Informatics and Health}.\hskip 1em plus 0.5em minus 0.4em\relax
  Springer, 2015, pp. 105--114.

\bibitem{kuo2015unified}
C.-T. Kuo, X.~Wang, P.~Walker, O.~Carmichael, J.~Ye, and I.~Davidson, ``Unified
  and contrasting cuts in multiple graphs: application to medical imaging
  segmentation,'' in \emph{SIGKDD}, 2015.

\bibitem{cahill2016multiple}
N.~D. Cahill, H.~Singh, C.~Zhang, D.~A. Corcoran, A.~M. Prengaman, P.~S.
  Wenger, J.~F. Hamilton, P.~Bajorski, and A.~M. Michael, ``Multiple-view
  spectral clustering for group-wise functional community detection,''
  \emph{arXiv preprint arXiv:1611.06981}, 2016.

\end{thebibliography}
